\documentclass[11pt]{article}
\usepackage[preprint]{acl}

\usepackage{times}
\usepackage{latexsym}
\usepackage[T1]{fontenc}
\usepackage[utf8]{inputenc}
\usepackage{microtype}
\usepackage{inconsolata}

\usepackage{multirow}
\usepackage{siunitx}
\usepackage{pgfplots}
\usepackage{pgfplotstable}
\pgfplotsset{compat=1.17}
\usetikzlibrary{positioning}
\usepackage{booktabs}
\usepackage{graphicx}
\usepackage{amsmath}
\usepackage{amssymb}
\usepackage{xcolor}
\usepackage{caption}   
\usepackage{enumitem}  
\setlist{nosep,leftmargin=*}  

\title{When Audio-LLMs Don't Listen:\\A Cross-Linguistic Study of Modality Arbitration}

\author{
  Jayadev Billa\thanks{Unaffiliated researcher - previously at ISI@USC, Yahoo, Nuance, BBN.}\\
  San Jose, CA, USA\\
  \texttt{jbilla2004@gmail.com}
}

\begin{document}

\maketitle

\begin{abstract}
When audio and text conflict, speech-enabled language models follow text far more often than they do when arbitrating between two conflicting text sources, even under explicit instructions to trust the audio. We introduce ALME (Audio-LLM Modality Evaluation), a dataset of 57,602 controlled audio-text conflict stimuli across eight languages, together with Text Dominance Ratio (TDR), which measures how often a model follows conflicting text when instructed to follow audio. Gemini 2.0 Flash and GPT-4o show TDR 10--26$\times$ higher than a baseline that replaces audio with its transcript under otherwise identical conditions (Gemini 2.0 Flash: 16.6\% vs.\ 1.6\%; GPT-4o: 23.2\% vs.\ 0.9\%). These results suggest that text dominance reflects not only information content, but also an asymmetry in \emph{arbitration accessibility}, i.e., how easily the model can use competing representations at decision time. Framing the transcript as deliberately corrupted reduces TDR by 80\%, whereas forcing explicit transcription increases it by 14\%. A fine-tuning ablation further suggests that arbitration behavior depends more on LLM reasoning than on the audio input path alone. Across four audio-LLMs, we observe the same qualitative pattern with substantial cross-model and cross-linguistic variation.
\end{abstract}

\section{Introduction}

Speech-enabled LLMs are typically deployed with additional text context, including system prompts, conversation history, and retrieved documents. If the text context conflicts with the user’s speech, the model makes a decision as to which source to follow. Existing evaluations largely measure transcription accuracy or downstream task performance under aligned inputs \citep{wang2024audiobench, conneau2023fleurs}, implicitly assuming that once speech is understood, downstream reasoning is unaffected by whether the input arrived as audio or text. Here we ask a simple question: \emph{when spoken audio and text conflict, which modality does the model follow?}

Such conflicts arise naturally in deployment. A user corrects a misunderstanding in speech, but the conversation history preserves the earlier error; a retrieved record conflicts with what the user just said aloud. In these settings, systematically favoring stale or incorrect text over the user’s actual words undermines the reliability of speech interfaces.

To study this problem, we introduce ALME and the Text Dominance Ratio (TDR), which measures how often a model follows conflicting text over audio under explicit instructions to follow what it hears. Our central finding is that audio-LLMs show a consistent preference to follow conflicting text over audio, even under explicit instructions to follow the audio. For Gemini 2.0 Flash, TDR rises from 1.6\% when arbitrating between two conflicting text sources to 16.6\% (10$\times$) when one source is presented as audio under otherwise identical framing. GPT-4o shows an even larger gap, from 0.9\% in the cascade baseline to 23.2\% (26$\times$) in the audio-text conflict condition. These results suggest that text dominance is not primarily a matter of \textbf{information content}, but of how easily the model can reason over competing representations when they conflict, i.e., \textbf{arbitration accessibility}.

We evaluate this through ALME, a benchmark spanning eight typologically diverse languages and four frontier audio-LLMs. Our contributions are as follows:
\begin{enumerate}
    \item \emph{ALME and TDR}: We introduce ALME, a benchmark for testing whether an audio-LLM follows spoken audio or conflicting text when instructed to follow the audio, together with Text Dominance Ratio (TDR) as the primary metric.
    \item \emph{Audio--text arbitration gap}: In both commercial and open-source models, we find a persistent preference to following conflicting text over audio despite explicit instructions to follow the audio. The strength of this preference varies widely across models and languages and can not be explained by transcript quality alone.
    \item \emph{Prompt-based mitigation}: Prompts that frame the conflicting transcript as unreliable reduces TDR by 80\%, whereas prompts that force explicit transcription before reasoning increases it by 14\%.
    \item \emph{Fine-tuning ablation}: Adapter-only tuning increases TDR, whereas LLM LoRA reduces it, suggesting that arbitration behavior depends more on the LLM’s conflict-resolution dynamics than on the audio adapter alone.
\end{enumerate}

\section{Related Work}

\subsection{Audio-LLM Evaluation and Audio-Text Conflict}

Several benchmarks evaluate audio-LLM capabilities under aligned inputs: AudioBench \citep{wang2024audiobench} covers speech recognition, audio QA, and captioning; AIR-Bench \citep{yang2024airbench} tests generative comprehension; MMAU \citep{sakshi2024mmau} focuses on expert-level audio reasoning; Dynamic-SUPERB \citep{huang2024dynamicsuperb} benchmarks instruction-following. These measure task accuracy but do not probe modality arbitration under conflict.

MCR-BENCH \citep{wang2025mcrbench} provides the first systematic evaluation of audio-text conflict, pairing audio with faithful, adversarial, or irrelevant text descriptions. Their primary metric, Text Influence Rate (TIR), measures how often adding text changes the model's answer relative to an audio-only baseline, with adversarial TIR reaching 98\% in some model--task conditions. TIR and our TDR measure different quantities: TIR captures whether text \emph{changes} the answer (a delta from no-text to text-present), while TDR measures which modality the model \emph{follows} when both are present and conflict (a proportion within conflict trials). The two are complementary: high TIR indicates text has influence; high TDR indicates how that influence resolves under explicit instructions to follow audio. LISTEN \citep{chen2025listen} measures lexical vs.\ acoustic emotion cue reliance, finding models inherit a ``preference for lexical cues.'' \citet{correa2025emis} demonstrate ``semantic anchoring'' in speech emotion recognition.

Our work addresses a different and complementary aspect of the problem (Table~\ref{tab:comparison}). Where prior work studies environmental audio or emotion, we focus on \emph{lexical speech content}. We evaluate across 8 typologically diverse languages, use controlled single-element semantic flips for precise attribution, and introduce a cascade baseline that separates modality-specific text dominance from general instruction-following limitations.

\begin{table}[t]
\centering
\caption{Comparison with related audio-text conflict benchmarks. ALME uniquely combines multilingual coverage, controlled semantic flips, and a cascade baseline.}
\label{tab:comparison}
\small
\resizebox{\columnwidth}{!}{%
\begin{tabular}{llll}
\toprule
\textbf{Dimension} & \textbf{MCR-BENCH} & \textbf{LISTEN} & \textbf{This work} \\
\midrule
Audio domain & Environmental & Emotion & Speech content \\
Languages & English & English & 8 languages \\
Conflict type & GPT desc. & Lex./acoustic & Semantic flips \\
Scale & 3,000 & 7,939 & 57,602 \\
Cascade baseline & No & No & Yes \\
FT ablation & No & No & Yes \\
\bottomrule
\end{tabular}}
\end{table}

\subsection{Knowledge Conflicts and Modality Bias}

The knowledge-conflict literature provides the closest conceptual framework for our work. Prior work studies how LLMs resolve conflicts between contextual evidence and parametric knowledge, identifying context-memory conflict as a recurring failure mode \citep{xu2024knowledge}. \citet{longpre2021entity} showed that models can be swayed by counterfactual context with substituted entities, a setup our semantic-flip construction partially echoes. \citet{xie2024adaptive} further showed that models are highly responsive to external evidence, but that this response depends on how that evidence aligns with existing priors. Our setting is related but not identical: rather than contextual evidence versus parametric memory, we study conflict between two external sources presented at inference time, one spoken and one textual. The shared question is how the model arbitrates between competing evidence sources.

A related pattern appears in vision-language models, where textual priors often override non-text evidence. Prior work has documented hallucination driven by language priors \citep{li2023pope, guan2024hallusionbench} and, more directly, persistent cross-modality conflict in which non-text evidence is underweighted relative to text \citep{zhu2024unraveling, zheng2025modality}. Our work brings this question into the audio-text setting and evaluates it with a controlled comparison to text-text conflict under matched reliability cues.

Our setting also parallels source-faithfulness problems in retrieval-augmented generation: in both cases, the model receives evidence that may conflict with text-derived expectations and must decide which source to follow. \citet{shi2023distracted} showed that LLMs are easily distracted by irrelevant context even in simple reasoning tasks, a finding our audio-text conflict results echo in the multimodal setting. From this perspective, TDR plays a role analogous to a source-faithfulness measure, quantifying how often the model fails to follow the primary evidence source under conflict. Methodologically, our cascade baseline adds an important control absent from prior work in both the audio-text and knowledge-conflict literatures. By comparing audio-text arbitration to text-text arbitration under identical reliability cues, we isolate the modality-specific component of text dominance from general instruction-following limitations.

\section{Methodology}

\subsection{Experimental Design}

We evaluate modality arbitration through controlled semantic conflict. Each stimulus presents a model with (1) natural speech from Common Voice \citep{ardila2020commonvoice}, (2) a modified transcript in which exactly one semantic element has been changed, and (3) a binary forced-choice question targeting that element. If the model follows the audio, it answers correctly; if it follows the transcript, it selects the conflicting answer. We quantify this behavior with \textbf{Text Dominance Ratio (TDR)}:
\[
\text{TDR} = \frac{\text{followed\_text}}{\text{followed\_text} + \text{followed\_audio}}
\]
TDR = 0 indicates perfect audio following; TDR = 1 indicates complete text dominance. Trials where the model produces an invalid response are excluded.

\subsection{Evaluation Conditions}

We evaluate four conditions, each serving a distinct diagnostic purpose (Figure~\ref{fig:conditions}):

\textbf{Audio-only}: Models receive only the audio and question, establishing whether the model can extract semantic content from speech alone.

\textbf{Text-only}: Models receive only the transcript and question, validating that questions are unambiguous.

\textbf{Aligned}: Models receive both audio and a correct transcript, establishing a multimodal ceiling.

\textbf{Conflict}: This is our primary evaluation condition. Audio and text disagree on exactly one semantic element. The prompt explicitly warns that ``the transcript may be incorrect'' and instructs the model to ``answer based on what you HEAR.''
\begin{figure}[t]
\centering
\resizebox{\columnwidth}{!}{%
\begin{tikzpicture}[
    modelbox/.style={rectangle, draw, fill=gray!20, rounded corners, minimum width=1.0cm, minimum height=0.6cm, font=\tiny\bfseries, align=center},
    inputbox/.style={rectangle, draw, fill=blue!10, rounded corners, font=\tiny, align=left, text width=2.9cm},
    outputbox/.style={rectangle, draw, fill=blue!5, rounded corners, font=\tiny, align=left},
    arrow/.style={->, >=stealth, thick},
    label/.style={font=\tiny\bfseries}
]

\node[label] at (-4.8, 0) {Audio-only};
\node[inputbox] (in1) at (-2.3, 0) {
\textbf{Audio:} ``...at \textit{three} o'clock''\\[0pt]
\textbf{Text:} [none]\\[0pt]
\textbf{Q:} three or five?
};
\node[modelbox] (m1) at (0.7, 0) {LLM};
\node[outputbox] (out1) at (2.5, 0) {\tiny ``three'' \checkmark};
\draw[arrow] (in1) -- (m1);
\draw[arrow] (m1) -- (out1);

\node[label, orange!80!black] at (-4.8, -1.3) {Conflict};
\node[inputbox, fill=orange!15] (in4) at (-2.3, -1.3) {
\textbf{Audio:} ``...at \textit{three} o'clock''\\[0pt]
\textbf{Text:} ``...at \textit{five} o'clock''\\[0pt]
\textbf{Q:} three or five?
};
\node[modelbox] (m4) at (0.7, -1.3) {LLM};
\node[outputbox, fill=yellow!20] (out4) at (2.5, -1.3) {\tiny ``five'' $\to$ TDR};
\draw[arrow] (in4) -- (m4);
\draw[arrow] (m4) -- (out4);

\end{tikzpicture}}
\caption{Audio-only (top) and conflict (bottom) conditions. In conflict, audio and text disagree; the model's forced-choice answer reveals modality preference.}
\label{fig:conditions}
\end{figure}

\subsection{Stimulus Generation and Dataset}

We source audio from Common Voice v22.0, selecting clips of 1.5--8 seconds with 3--25 word transcripts and positive community ratings. Rule-based matchers identify transcripts containing flippable semantic elements across four types: \textbf{number swaps} (e.g., ``three cats'' $\to$ ``five cats''), \textbf{negation insertion/removal} (``is open'' $\leftrightarrow$ ``is not open''), \textbf{adjective swaps} (``big house'' $\to$ ``small house''), and \textbf{time expression swaps} (``morning'' $\to$ ``evening''). Each per-language dictionary contains 6--65 entries per category (e.g., 30--52 for numbers, 23--64 for adjectives), adapted to the language's morphological patterns.

For each candidate, GPT-4.1-mini generates a neutral binary question targeting the flipped element. Questions must (a) target exactly the modified element, (b) have unambiguous answers, and (c) randomize answer order to avoid position bias. All stimuli then undergo an additional LLM-based review pass for consistency and answerability. The final dataset comprises 57,602 stimuli across eight languages spanning four scripts and five language families (Table~\ref{tab:dataset}). Approximately 4,400 stimuli share the same (language, reference text, flip type) tuple with different audio recordings from different speakers; the number of unique linguistic conflict scenarios is thus $\sim$53,000.

\begin{table}[t]
\centering
\caption{ALME dataset statistics by language ($n{=}57{,}602$).}
\label{tab:dataset}
\small
\begin{tabular}{lrll}
\toprule
\textbf{Lang} & \textbf{$n$} & \textbf{Script} & \textbf{Family} \\
\midrule
EN & 7,328 & Latin & Germanic \\
DE & 7,210 & Latin & Germanic \\
FR & 7,413 & Latin & Romance \\
IT & 7,296 & Latin & Romance \\
PT & 7,237 & Latin & Romance \\
AR & 6,922 & Arabic & Semitic \\
JA & 7,041 & Kana/Kanji & Japonic \\
ZH & 7,155 & Hanzi & Sinitic \\
\midrule
\textbf{Total} & \textbf{57,602} & \multicolumn{2}{c}{4 scripts, 5 families} \\
\bottomrule
\end{tabular}
\end{table}

\subsection{Stimulus Validation}

The forced-choice format makes the evaluation less sensitive to minor wording issues. Because models choose between two semantically distinct options (e.g., ``three'' vs.\ ``five''), small phrasing variations do not usually affect discriminative validity. Native speakers reviewed 40 randomly sampled stimuli each for English, Japanese, and Portuguese. Under a forced-choice discriminability criterion (a stimulus is valid if and only if the two choices map unambiguously to different modalities), 97.5\% of English, 92.5\% of Japanese, and 90\% of Portuguese stimuli were judged valid.

\subsection{Models Evaluated}

We evaluate four audio-LLMs: \textbf{GPT-4o-audio} \citep{openai2024gpt4o}, \textbf{Gemini 2.0 Flash} \citep{team2024gemini}, \textbf{Ultravox v0.6-llama-3.1-8b} \citep{ultravox2024}, and \textbf{Qwen2-Audio-7B-Instruct} \citep{chu2024qwen2audio}. Models receive a system prompt enforcing structured JSON output, audio (WAV, 16kHz mono), and a user message with the transcript, question, and answer choices (Appendix~\ref{app:prompts}). Responses are parsed deterministically via exact string match; parse failure rates are $<$1\% for all models (Appendix~\ref{app:errors}). All evaluations use temperature $= 0$ for reproducibility. Throughout this paper, references to GPT-4o, Gemini, Ultravox, and Qwen2 refer to these specific model versions unless otherwise noted.

\subsection{Cascade Baseline}
\label{sec:cascade}

To determine whether text dominance is specific to multimodal processing, we construct cascade baselines for Gemini and GPT-4o: Whisper large-v3 \citep{radford2023whisper} transcribes the audio, then each model (text-only mode) answers with two competing transcripts and explicit reliability cues. We evaluate three cascade conditions: \textbf{Cascade-A} (ASR transcript only, establishing baseline accuracy), \textbf{Cascade-B} (ASR transcript + conflicting text, filtered to stimuli where Whisper correctly transcribed the key element; $n{=}53{,}747$; Appendix~\ref{app:whisper}), and \textbf{Cascade-C} (reference transcript + conflicting text, no ASR noise). If cascade TDR matches multimodal TDR, text dominance is a general instruction-following failure; if much lower, it is multimodal-specific.

\subsection{Prompt Intervention}

Using Gemini on 14,369 stimuli from English and Japanese, representing a European and a CJK language with contrasting TDR levels, we compare four prompt variants (Appendix~\ref{app:intervention}): \textbf{baseline} (``may contain errors''), \textbf{adversarial} (``deliberately corrupted''), \textbf{audio-first} (requires explicit transcription before answering), and \textbf{explicit-ignore} (``COMPLETELY IGNORE the transcript'').

\subsection{TTS Resynthesis}

To probe whether audio generation method affects modality arbitration, we resynthesized all 57,602 stimuli using Azure Neural TTS \citep{azuretts2024}. Synthetic speech reduces speaker-specific variation, accent diversity, and natural disfluencies while preserving lexical content. For each language, we selected six neural voices with balanced gender distribution.

\subsection{Fine-Tuning Ablation}

We chose Ultravox for the ablation because its architecture cleanly separates the audio projection layer from the LLM, allowing us to isolate each component's contribution to arbitration behavior. We fine-tuned under two conditions: (1) \textbf{adapter-only}, updating only the multi-modal projection layer (LLM and audio encoder frozen); and (2) \textbf{LLM LoRA} (rank 16, $\alpha{=}32$) on the LLM's attention projections (adapter and encoder frozen). Both were trained on conflict stimuli from two flip types (number swaps, adjective swaps) with a 50/50 mix of conflict and aligned examples, and evaluated on \emph{held-out} flip types ($n{=}29{,}263$).

\subsection{Statistical Analysis}

We report Wilson score 95\% confidence intervals for all proportions. Because multiple trials may come from the same speaker (20,444 unique speakers), we also compute speaker-clustered bootstrap CIs (1,000 iterations).\footnote{The bootstrap script computes TDR over all conflict trials (including ``other'' responses in the denominator), while pooled TDR excludes them. With $<$1\% parse failures, this produces at most $\sim$0.6pp discrepancy; all reported values use the pooled definition.} Cohen's $h$ quantifies effect magnitude. For pairwise language comparisons, we apply Bonferroni correction ($\alpha_{\text{adj}} = 0.0018$).

\section{Results}
\label{sec:results}

\subsection{The Arbitration Gap}
\label{sec:10x_gap}

Both Gemini and GPT-4o achieve high accuracy in control conditions (97--98\% and 93--99\%, respectively), confirming strong comprehension of both modalities. Under audio-text conflict, however, both models follow the conflicting text far more often than expected: Gemini TDR is 16.6\% and GPT-4o TDR is 23.2\%. Table~\ref{tab:cascade} shows that this text preference is modality-specific. When we replace audio with its Whisper transcript and present two competing text sources with identical reliability cues, cascade TDR drops to 1.6\% for Gemini and 0.9\% for GPT-4o. The gap between multimodal and cascade TDR is large for both models (Cohen's $h{=}0.59$ for Gemini, $h{=}0.82$ for GPT-4o), confirming that text dominance is specific to audio-text processing, not a general instruction-following failure. Notably, GPT-4o is \emph{better} than Gemini at text-text arbitration (0.9\% vs.\ 1.6\%) yet \emph{worse} at audio-text arbitration (23.2\% vs.\ 16.6\%), indicating that text-text and audio-text arbitration are independent capabilities. Both cascade baselines show minimal position effect ($\leq$0.3\%).

\begin{table}[t]
\centering
\caption{Multimodal vs.\ cascade TDR for two models. Both show a large gap, confirming text dominance is modality-specific. GPT-4o is better at text-text arbitration yet worse at audio-text.}
\label{tab:cascade}
\small
\resizebox{\columnwidth}{!}{%
\begin{tabular}{llcc}
\toprule
\textbf{Condition} & \textbf{Input} & \textbf{Gemini} & \textbf{GPT-4o} \\
\midrule
Multimodal & Audio + conflict text & 16.6\% & 23.2\% \\
\midrule
Cascade-B & ASR + conflict text & 1.6\% & 0.9\% \\
Cascade-C & Ref. + conflict text & 1.0\% & 0.7\% \\
\bottomrule
\end{tabular}}
\end{table}

One might ask whether text dominance simply reflects poor audio comprehension. For Gemini, this is clearly not the case: audio-only accuracy (97.2\%) exceeds Cascade-A accuracy (93.9\%), meaning the model extracts \emph{more} information from audio than from ASR transcripts (Appendix~\ref{app:audio_vs_asr}). For GPT-4o, the pattern reverses: Cascade-A (96.2\%) exceeds audio-only (92.7\%). Yet GPT-4o still exhibits substantial text dominance under conflict (23.2\% TDR vs.\ 0.9\% cascade). Text dominance persists whether audio comprehension is strong or weak, ruling out audio quality as the primary explanation.

\subsection{Four-Model Comparison}

The arbitration gap is not limited to Gemini and GPT-4o. All four models exhibit text dominance under conflict (Figure~\ref{fig:tdr_models}, Table~\ref{tab:tdr_overall}). What varies is the degree. Gemini (16.6\%) and GPT-4o (23.2\%) follow text in a minority of trials; Ultravox (48.8\%) follows text in roughly half of trials; Qwen2-Audio (63.2\%) follows text in nearly two-thirds of trials. The 47\% spread between Gemini and Qwen2 represents a qualitative behavioral difference (Cohen's $h{=}1.01$, very large effect). Lower TDR tends to coincide with higher audio-only accuracy, though we do not claim audio comprehension alone determines arbitration behavior.

\begin{figure}[t]
\centering
\begin{tikzpicture}
\begin{axis}[
    ybar,
    width=\columnwidth,
    height=4.5cm,
    bar width=0.6cm,
    ylabel={Text Dominance Ratio (\%)},
    ylabel style={font=\small},
    symbolic x coords={Gemini,GPT-4o,Ultravox,Qwen2},
    xtick=data,
    xticklabel style={font=\small},
    yticklabel style={font=\small},
    ymin=0, ymax=80,
    ytick={0,20,40,60,80},
    nodes near coords,
    nodes near coords style={font=\tiny},
    every node near coord/.append style={yshift=6pt},
    enlarge x limits=0.2,
    grid=major,
    grid style={dashed, gray!30},
]
\addplot[fill=teal!70, error bars/.cd, y dir=both, y explicit]
    coordinates {
        (Gemini,16.6) +- (0,0.3)
        (GPT-4o,23.2) +- (0,0.35)
        (Ultravox,48.8) +- (0,0.4)
        (Qwen2,63.2) +- (0,0.4)
    };
\draw[solid, violet!70!black, line width=1.5pt] (axis cs:Gemini,50) -- (axis cs:Qwen2,50);
\end{axis}
\node[font=\tiny, violet!70!black] at (1.5,2.7) {50\%};
\end{tikzpicture}
\caption{TDR across four audio-LLMs ($n{=}57{,}602$). Horizontal line: TDR${=}$50\% (text followed in half of conflict trials).}
\label{fig:tdr_models}
\end{figure}

\begin{table}[t]
\centering
\caption{Overall TDR (\%) with speaker-clustered bootstrap CIs and control condition accuracies ($n{=}57{,}602$, 20,444 speakers). TDR values are bootstrap CI midpoints; pooled TDR (used in running text) differs by up to 0.5pp.}
\label{tab:tdr_overall}
\small
\begin{tabular}{lcccc}
\toprule
\textbf{Model} & \textbf{TDR} & \textbf{AO} & \textbf{TO} & \textbf{Alig.} \\
\midrule
Gemini & 16.5 {\tiny [15.7, 17.5]} & 97.2 & 98.2 & 97.8 \\
GPT-4o & 23.1 {\tiny [22.3, 24.0]} & 92.7 & 98.1 & 98.6 \\
Ultravox & 48.7 {\tiny [48.2, 49.3]} & 82.2 & 83.1 & 86.8 \\
Qwen2 & 62.7 {\tiny [62.0, 63.4]} & 77.9 & 77.2 & 80.5 \\
\bottomrule
\end{tabular}
\end{table}

\subsection{Cross-Linguistic Variation}

Table~\ref{tab:tdr_language} reveals substantial cross-linguistic variation. Three of four models (Gemini, GPT-4o, Ultravox) show 2--4$\times$ higher TDR for CJK/Arabic than European languages. For Gemini, Chinese shows $\sim$4$\times$ higher TDR than English (31.8\% vs.\ 8.1\%); for GPT-4o, the ratio is $\sim$3.8$\times$ (42.8\% vs.\ 11.3\%). In practical terms, a model that follows audio 92\% of the time for English but only 68\% for Chinese is exhibiting qualitatively different arbitration behavior.

Qwen2-Audio shows a reversed pattern: CJK/Arabic languages have \emph{lower} TDR (50--56\%) than European languages (64--71\%), possibly reflecting differences in training data composition. All $\chi^2$ tests ($\text{df}{=}7$) yield $p < 0.001$.

\begin{table}[t]
\centering
\caption{TDR (\%) by language. Three models show higher TDR for CJK/Arabic; Qwen2-Audio reverses this pattern. All $\chi^2$ tests: $p{<}0.001$.}
\label{tab:tdr_language}
\small
\begin{tabular}{lcccc}
\toprule
\textbf{Lang} & \textbf{Gem.} & \textbf{GPT} & \textbf{Ultr.} & \textbf{Qwen2} \\
\midrule
EN & 8.1 & 11.3 & 39.2 & 69.2 \\
DE & 11.3 & 13.6 & 48.5 & 63.6 \\
FR & 12.6 & 19.8 & 47.7 & 71.0 \\
IT & 8.9 & 21.4 & 40.1 & 70.4 \\
PT & 10.3 & 21.9 & 49.4 & 71.5 \\
\midrule
AR & 19.9 & 25.1 & 53.8 & 50.4 \\
JA & 30.2 & 30.2 & 55.1 & 51.8 \\
ZH & 31.8 & 42.8 & 57.6 & 56.2 \\
\midrule
\textbf{All} & \textbf{16.6} & \textbf{23.2} & \textbf{48.8} & \textbf{63.2} \\
\bottomrule
\end{tabular}
\end{table}

\subsection{TTS Robustness}

Replacing natural speech with Azure Neural TTS produces different effects across model families: API models show stable or decreased TDR (GPT-4o $-$4.2\%, Gemini +0.2\% n.s.), while open-source models shift toward text (Ultravox +3.5\%, Qwen2 +2.2\%, both $p{<}0.001$). This may reflect a domain shift if open-source models were trained primarily on natural speech. Full per-language results are in Appendix~\ref{app:tts}.

\subsection{Prompt Intervention}

Table~\ref{tab:intervention} shows that prompt framing significantly affects TDR, but the direction depends on the intervention type.

\textbf{Reducing text trust lowers TDR.} Adversarial framing (``deliberately corrupted'') reduces TDR from 19\% to 3.8\%, an 80\% relative reduction (Cohen's $h{=}0.51$, medium effect), without any model modification. Explicit-ignore (``COMPLETELY IGNORE'') also helps (7.9\%), but less effectively ($h{=}0.33$). Epistemic framing outperforms behavioral commands, suggesting the model's arbitration is sensitive to source reliability attributions.

\textbf{Forcing explicit transcription increases TDR.} Audio-first (requiring explicit transcription before answering) \emph{increases} TDR to 33\%. Forcing transcription appears to increase reliance on the text representation rather than reducing it.

\begin{table}[t]
\centering
\caption{Prompt intervention TDR (\%) (Gemini, EN+JA, $n{=}14{,}369$). Adversarial framing substantially closes the cross-linguistic gap.}
\label{tab:intervention}
\small
\begin{tabular}{lccccc}
\toprule
\textbf{Variant} & \textbf{EN} & \textbf{JA} & \textbf{Overall} & \textbf{$h$} \\
\midrule
baseline & 8.1 & 30.2 & 19.0 & --- \\
adversarial & 2.0 & 5.7 & 3.8 & $-$0.51 \\
explicit-ignore & 3.8 & 12.3 & 7.9 & $-$0.33 \\
audio-first & 19.2 & 47.4 & 33.0 & +0.32 \\
\bottomrule
\end{tabular}
\end{table}

The intervention effects generalize across languages: adversarial framing reduces Japanese TDR from 30.2\% to 5.7\% ($-$24.5pp), substantially closing the cross-linguistic gap with English (5.7\% vs.\ 2.0\%). The 3.8\% adversarial floor vs.\ 1.6\% cascade baseline leaves a 2.2\% gap, possibly reflecting a residual cost of audio-text relative to text-text arbitration even under strong transcript distrust.

\subsection{Fine-Tuning Ablation}
\label{sec:ablation}

Table~\ref{tab:ablation} shows the two interventions produce opposite effects on held-out flip types ($n{=}29{,}263$). LLM LoRA nearly halves TDR (49.4\% $\to$ 25.5\%, $\Delta{=}-23.9$), while adapter-only fine-tuning sharply \emph{increases} it (49.4\% $\to$ 75.9\%, $\Delta{=}+26.5$). Both effects generalize to unseen flip types (Appendix~\ref{app:ablation}), suggesting a broader shift in arbitration behavior rather than memorization of specific conflict patterns.

\begin{table}[t]
\centering
\caption{Ultravox fine-tuning ablation on held-out flip types ($n{=}29{,}263$). Adapter-only increases TDR; LLM LoRA reduces it.}
\label{tab:ablation}
\small
\resizebox{\columnwidth}{!}{%
\begin{tabular}{lccc}
\toprule
\textbf{Condition} & \textbf{TDR (\%)} & \textbf{AO Acc} & \textbf{Alig.\ Acc} \\
\midrule
Baseline & 49.4 & 80.1 & 85.3 \\
Adapter-only & 75.9 (+26.5) & 53.8 ($-$26.3) & 84.5 \\
LLM LoRA & 25.5 ($-$23.9) & 56.3 ($-$23.8) & 86.2 \\
\bottomrule
\end{tabular}}
\end{table}

While Ultravox's baseline TDR of $\sim$50\% could suggest the model is simply confused rather than arbitrating, the ablation results suggest otherwise: adapter-only and LLM LoRA training produce \emph{opposite} directional shifts from the same baseline (+26.5 vs.\ $-$23.9), which would not occur if the model were responding randomly. The adapter result is counterintuitive: training the projection layer on conflict stimuli, where ground truth favors audio, made the model \emph{more} text-dominant. One interpretation is that the adapter produces embeddings that diverge from representations the frozen LLM was trained to process, causing it to default more strongly to text. LLM LoRA, by contrast, appears to affect the arbitration process more directly, learning to upweight audio-derived content when conflict is detected.

Both conditions share an unexpected side effect: audio-only accuracy drops $\sim$25pp (to 54--56\%), while aligned accuracy is preserved (85--86\%). This suggests that conflict training improves the model's ability to choose between modalities, but at the cost of degrading its ability to comprehend audio in isolation.

LLM LoRA also reveals a cross-linguistic pattern: European languages approach Gemini-level TDR (EN: 37\% $\to$ 7.6\%, DE: 49\% $\to$ 10.0\%, IT: 44\% $\to$ 9.1\%), while CJK/Arabic remain at 49--54\% TDR (Appendix~\ref{app:ablation}). This mirrors the gradient in the main experiment and suggests LoRA intervention is more effective in languages where audio-text alignment is easier for the model to exploit.

\section{Discussion}

\subsection{Why Do Models Prefer Text Under Conflict?}
\label{sec:discussion_framework}

All four models can comprehend audio well enough to answer questions correctly when no conflicting text is present (77--97\% audio-only accuracy). Yet under conflict, every model follows text at least some of the time, and all do so far more than in the text-text cascade baseline. Why?

We propose that text dominance reflects not a failure of comprehension but a failure of \emph{arbitration}: models can extract information from audio, but cannot weigh it against competing text as effectively as they weigh text against text. We call this property \emph{arbitration accessibility}. Four observations support this interpretation:

\textbf{Text dominance is independent of audio quality.} If text dominance were caused by poor audio comprehension, models with better audio should show less text dominance. Gemini's audio-only accuracy (97.2\%) exceeds its cascade accuracy (93.9\%), yet its multimodal TDR is still 16.6\%. GPT-4o's audio is \emph{worse} than Whisper ASR (92.7\% vs.\ 96.2\%), yet it shows an even larger multimodal-to-cascade gap (23.2\% vs.\ 0.9\%). The problem is not that models cannot hear; it is that they cannot use what they hear when text disagrees.

\textbf{Forcing transcription makes things worse.} If the bottleneck were audio comprehension, explicitly requiring audio transcription before answering should help. Instead, TDR increases from 19\% to 33\%. The model already relies on text representations during arbitration; explicit transcription reinforces that reliance rather than providing a new information source.

\textbf{Reducing text trust works; improving audio processing does not.} Adversarial framing (``this text is deliberately corrupted'') reduces TDR by 80\%, while explicit-ignore instructions (``COMPLETELY IGNORE the transcript'') are less effective ($-$58\%). Changing the model's \emph{trust} in text is more effective than changing how it \emph{processes} audio, consistent with arbitration being the bottleneck.

\textbf{Fine-tuning the LLM helps; fine-tuning the adapter hurts.} Training the audio projection layer on conflict data increases TDR (+26.5\%), while LoRA on the LLM decreases it ($-$23.9\%). If text dominance resided in audio encoding, adapter training should help. Instead, only modifying the LLM's conflict-resolution behavior reduces text dominance, consistent with arbitration accessibility being a property of the LLM rather than the encoder.

Similar patterns have been observed in vision-language models \citep{li2023pope, zhu2024unraveling}, where text priors override visual evidence. The specificity of our effect to audio-text (rather than text-text) arbitration suggests that the issue is not language priors in general, but how easily non-text representations can participate in the model's reasoning process. We note that training data composition (predominantly text) is a plausible contributing factor that our experiments cannot fully disentangle from architectural effects; the two likely interact.

\subsection{Cross-Model Variation}

The 47\% TDR spread across models shows that text dominance varies substantially, likely reflecting differences in architecture and training. All models receive identical conflict instructions, yet Gemini follows text 16.6\% of the time while Qwen2 does so 63.2\% of the time. Ultravox follows text in roughly half of conflict trials, and Qwen2 in nearly two-thirds. Lower TDR correlates with higher audio-only accuracy across models, suggesting that audio comprehension quality may be one contributing factor, though the cascade results show it is not the only one.

\subsection{Prompt-Based Mitigation}

The prompt intervention results (Section~4.5) point to a practical asymmetry: reducing the model's trust in the transcript is effective, but instructing the model to attend more carefully to audio is not. Adversarial framing achieves 3.8\% TDR, approaching but not reaching the cascade baseline (1.6\% for Gemini). The remaining 2.2\% gap may represent a residual cost of audio-text arbitration that cannot be closed by prompt design alone.

\subsection{Cross-Linguistic Implications}

The 2--4$\times$ TDR gap between European and CJK/Arabic languages (Section~4.3) has practical consequences: speakers of CJK and Arabic languages may experience systematically lower audio fidelity in speech-based interactions. While part of this variation tracks Whisper ASR accuracy (Appendix~\ref{app:whisper}), the cascade baseline shows it cannot be the full explanation, since cascade TDR remains below 2\% for all languages. These cross-linguistic disparities connect to broader documented ASR performance gaps \citep{koenecke2020racial, meyer2020artie} and warrant attention from practitioners deploying multilingual speech systems.

\subsection{Implications for System Design}

Our findings have direct implications for building reliable audio-LLM systems.

\textbf{Model selection encodes an arbitration policy.} The 47\% TDR spread suggests that choosing Qwen2-Audio vs.\ GPT-4o implicitly chooses how the system handles conflicting inputs. For applications where audio should be authoritative (voice interfaces, transcription correction), models with lower TDR are preferable.

\textbf{Multilingual deployment requires per-language evaluation.} A model's aggregate TDR masks substantial cross-linguistic variation. Even Gemini, the lowest-TDR model, shows 4$\times$ higher TDR for Chinese (31.8\%) than English (8.1\%). English-only benchmarks are insufficient for predicting multilingual behavior.

\textbf{Epistemic framing is a practical mitigation.} When audio fidelity matters, framing accompanying text as potentially corrupted reduces TDR by 80\% without model modification. This is immediately deployable in system prompts.

\textbf{Architectural interventions have asymmetric effects.} Fine-tuning the audio adapter increases text dominance, while fine-tuning the LLM via LoRA reduces it, though at the cost of $\sim$25pp degradation in standalone audio comprehension. Resolving this trade-off is an open problem.

\section{Conclusion}

We introduced ALME, a benchmark for making modality arbitration observable through controlled audio-text conflict across eight languages and four audio-LLMs. Across models, we find a large gap between audio-text and text-text arbitration: Gemini follows conflicting text 16.6\% of the time under audio-text conflict versus 1.6\% under text-text conflict, while GPT-4o shows an even larger gap (23.2\% vs.\ 0.9\%). These results suggest that text dominance is not explained by information quality alone, but also by how easily different representations can be used during conflict resolution.

This perspective helps explain the intervention and ablation results. Reducing trust in the transcript lowers TDR substantially, whereas forcing explicit transcription before answering increases it. Fine-tuning the audio adapter increases TDR, whereas LLM LoRA reduces it, consistent with arbitration behavior depending more on the LLM’s conflict-resolution dynamics than on the audio encoder alone.

The practical implication is that evaluating speech-enabled LLMs requires more than transcription accuracy or downstream task performance. It also requires measuring how models resolve conflicts between spoken and textual evidence, especially in multilingual settings where arbitration behavior can vary substantially across languages. Our evaluation dataset and code are publicly available.\footnote{\url{https://github.com/jb1999/alme-benchmark}}

\section*{Limitations}

Our forced-choice format makes modality arbitration directly observable but does not capture open-ended generation, where reliance on audio or text is harder to disentangle. Extending arbitration analysis to free-form interaction will require new attribution methods. Our flip types (numbers, negations, adjectives, times) cover common semantic dimensions but not prosodic conflicts (sarcasm, emphasis) or non-speech audio.

We focus on TDR, measuring how often models follow text when instructed to follow audio. A complementary metric, Audio Dominance Ratio, measuring how often models follow audio when instructed to follow text, could reveal additional patterns. We leave this symmetric analysis for future work.

As the prompt intervention experiment (Section~4.5) shows, TDR is sensitive to how the conflict transcript is framed, varying from 3.8\% under adversarial framing to 33\% under audio-first prompting. We did not conduct a systematic stability analysis with minor rephrasings of the baseline conflict prompt, though the consistency of TDR patterns across four models, eight languages, and five flip types suggests that the observed effects reflect genuine model behavior rather than prompt-specific artifacts.

The fine-tuning ablation was conducted on Ultravox only due to resource constraints. Both fine-tuned conditions degraded audio-only accuracy by $\sim$25\%, suggesting a comprehension--arbitration trade-off that warrants further investigation with larger training sets.

Human validation was conducted by a single native speaker per language on 40 items each (EN, JA, PT). The remaining five languages (AR, DE, FR, IT, ZH) rely on indirect evidence of stimulus quality: the consistent cross-model TDR patterns across languages and flip types would be unlikely to emerge from systematically flawed stimuli. While the forced-choice format makes validity assessment near-deterministic, inter-annotator agreement was not computed. Stimuli derive from Common Voice, which all models likely encountered during training; this affects all models equally, preserving cross-model comparisons.

Our results depend on specific model versions evaluated in early 2026. All evaluations use temperature $= 0$; stochastic sampling may yield different distributions.

\section*{Ethical Considerations}

This study uses publicly available speech data from Mozilla Common Voice (CC-0 license). No new human subjects data was collected; human validation used the authors' assessments of existing stimuli.

Our findings reveal that audio-LLMs may override spoken input for certain language groups (2--4$\times$ higher text dominance for CJK/Arabic in most models). While we do not measure population-level fairness, these disparities warrant attention from practitioners deploying multilingual speech systems.

The adversarial prompt intervention demonstrates that text dominance can be both reduced (adversarial framing) and increased (audio-first prompting). We report these findings to inform defensive prompt design; the techniques could theoretically be misused to manipulate model behavior, though they require control over the system prompt.


\bibliography{references}

\appendix


\section{Prompt Templates}
\label{app:prompts}

\noindent
\small
\fbox{\parbox{0.95\columnwidth}{
\colorbox{gray!15}{\parbox{0.88\columnwidth}{\textbf{System:} You are an expert speech and language analyst.\\[2pt]
RESPONSE FORMAT: Output ONLY valid JSON: \{``answer'': ``\textless choice\textgreater'', ``confidence'': 0.0--1.0, ``rationale'': ``brief''\}. Your ``answer'' MUST be EXACTLY one of the provided choices, copied verbatim. Use the SAME language and script as the choices.}}\\[4pt]
\textbf{Aligned condition:}\\[2pt]
\colorbox{cyan!25}{\parbox{0.88\columnwidth}{Audio: [attached]\\Transcript: ``\textit{\textless transcript\textgreater}''\\QUESTION: \textit{\textless question\textgreater}\\CHOICES: [\textit{A}, \textit{B}]\\Your answer MUST be exactly one of the CHOICES above.}}\\[4pt]
\textbf{Conflict condition:}\\[2pt]
\colorbox{orange!30}{\parbox{0.88\columnwidth}{Audio: [attached]\\Transcript (may contain errors): ``\textit{\textless transcript\textgreater}''\\IMPORTANT: The transcript may be incorrect. Answer based on what you HEAR in the audio.\\QUESTION: \textit{\textless question\textgreater}\\CHOICES: [\textit{A}, \textit{B}]\\Your answer MUST be exactly one of the CHOICES above.}}
}}
\captionof{figure}{Prompt templates for aligned and conflict conditions.}
\label{fig:prompts}
\normalsize

\section{Prompt Intervention Variants}
\label{app:intervention}

\noindent
\small
\fbox{\parbox{0.95\columnwidth}{
\textbf{Baseline:}\\[2pt]
\colorbox{gray!15}{\parbox{0.88\columnwidth}{Transcript (may contain errors): ``\textit{<text>}''\\IMPORTANT: The transcript may be incorrect. Answer based on what you HEAR.}}\\[4pt]
\textbf{Adversarial:}\\[2pt]
\colorbox{red!15}{\parbox{0.88\columnwidth}{Transcript (DELIBERATELY CORRUPTED): ``\textit{<text>}''\\WARNING: The transcript has been intentionally altered.}}\\[4pt]
\textbf{Audio-first:}\\[2pt]
\colorbox{cyan!15}{\parbox{0.88\columnwidth}{Transcript (may contain errors): ``\textit{<text>}''\\INSTRUCTIONS: 1. Transcribe what you hear. 2. Answer based on your transcription.}}\\[4pt]
\textbf{Explicit-ignore:}\\[2pt]
\colorbox{orange!20}{\parbox{0.88\columnwidth}{Transcript (UNRELIABLE - IGNORE): ``\textit{<text>}''\\CRITICAL: COMPLETELY IGNORE the transcript.}}
}}
\captionof{figure}{Prompt intervention variants.}
\label{fig:intervention_prompts}
\normalsize

\section{Whisper Key Element Accuracy}
\label{app:whisper}

\begin{center}\small
\captionof{table}{Whisper large-v3 key element accuracy by language.}
\label{tab:whisper_accuracy}
\begin{tabular}{lcr}
\toprule
\textbf{Lang} & \textbf{Accuracy} & \textbf{$n$ (correct)} \\
\midrule
EN & 96.1\% & 7,044 \\
DE & 97.7\% & 7,046 \\
FR & 91.3\% & 6,766 \\
IT & 97.9\% & 7,144 \\
PT & 97.5\% & 7,059 \\
AR & 89.2\% & 6,177 \\
JA & 91.5\% & 6,446 \\
ZH & 84.8\% & 6,065 \\
\midrule
\textbf{All} & \textbf{93.3\%} & \textbf{53,747} \\
\bottomrule
\end{tabular}
\end{center}

\section{Audio-Only vs.\ Cascade Accuracy}
\label{app:audio_vs_asr}

\begin{center}\small
\captionof{table}{Audio-only vs.\ Cascade-A accuracy (\%) by language. For Gemini, audio outperforms ASR across all languages. For GPT-4o, the pattern reverses: Cascade-A exceeds audio-only overall.}
\label{tab:audio_vs_asr}
\resizebox{\columnwidth}{!}{%
\begin{tabular}{l rrr rrr}
\toprule
 & \multicolumn{3}{c}{\textbf{Gemini}} & \multicolumn{3}{c}{\textbf{GPT-4o}} \\
\cmidrule(lr){2-4} \cmidrule(lr){5-7}
\textbf{Lang} & \textbf{AO} & \textbf{Casc.} & \textbf{$\Delta$} & \textbf{AO} & \textbf{Casc.} & \textbf{$\Delta$} \\
\midrule
EN & 97.7 & 96.7 & +1.0 & 96.3 & 96.3 & +0.0 \\
DE & 98.2 & 97.8 & +0.4 & 96.3 & 97.8 & $-$1.5 \\
FR & 96.5 & 95.8 & +0.6 & 95.7 & 97.1 & $-$1.4 \\
IT & 98.8 & 98.4 & +0.4 & 96.2 & 98.5 & $-$2.3 \\
PT & 98.0 & 97.7 & +0.3 & 91.0 & 97.4 & $-$6.4 \\
AR & 95.7 & 83.0 & +12.7 & 88.1 & 94.9 & $-$6.8 \\
JA & 96.2 & 88.2 & +8.0 & 91.3 & 95.8 & $-$4.5 \\
ZH & 96.2 & 92.6 & +3.6 & 86.0 & 92.0 & $-$6.0 \\
\midrule
\textbf{All} & \textbf{97.2} & \textbf{93.9} & \textbf{+3.3} & \textbf{92.7} & \textbf{96.2} & \textbf{$-$3.5} \\
\bottomrule
\end{tabular}}
\end{center}

\section{Stimuli Distribution by Flip Type}
\label{app:flipdist}

\begin{center}\small
\captionof{table}{Stimuli count by semantic flip type.}
\label{tab:flipdist}
\begin{tabular}{lr}
\toprule
\textbf{Flip Type} & \textbf{Count} \\
\midrule
time\_swap & 14,557 \\
adjective\_swap & 14,200 \\
number\_swap & 14,139 \\
negation\_remove & 8,269 \\
negation\_add & 6,437 \\
\midrule
\textbf{Total} & \textbf{57,602} \\
\bottomrule
\end{tabular}
\end{center}

\section{TDR by Flip Type}
\label{app:tdr_fliptype}

\begin{center}\small
\captionof{table}{TDR (\%) by semantic flip type. Negation addition consistently produces the highest TDR across all models.}
\label{tab:tdr_fliptype}
\begin{tabular}{lcccc}
\toprule
\textbf{Flip Type} & \textbf{Gem.} & \textbf{GPT} & \textbf{Ultr.} & \textbf{Qwen2} \\
\midrule
adjective & 11.6 & 17.2 & 49.1 & 66.7 \\
number & 14.4 & 25.3 & 47.0 & 63.3 \\
time & 17.7 & 23.0 & 49.0 & 63.5 \\
neg\_remove & 17.0 & 19.5 & 43.0 & 51.6 \\
neg\_add & 29.2 & 37.0 & 59.5 & 69.4 \\
\bottomrule
\end{tabular}
\end{center}

\section{Position Bias}
\label{app:position}

\begin{center}\small
\captionof{table}{Position bias in forced-choice answer selection.}
\label{tab:position_bias}
\begin{tabular}{lrrrr}
\toprule
\textbf{Model} & \textbf{Pos 0} & \textbf{Pos 1} & \textbf{Gap} & \textbf{Severity} \\
\midrule
GPT-4o & 49.5\% & 50.5\% & 1.0\% & Negligible \\
Gemini & 48.2\% & 51.8\% & 1.2\% & Negligible \\
Ultravox & 51.6\% & 48.4\% & 2.5\% & Mild \\
Qwen2 & 36.4\% & 63.6\% & 27.6\% & Severe \\
\bottomrule
\end{tabular}
\end{center}

\medskip\noindent
Qwen2-Audio shows severe recency bias (27.6\% gap). Position-corrected TDR (62.6\%) matches overall TDR (62.7\%), confirming position bias does not inflate our TDR conclusions.

\section{Prompt Intervention by Flip Type}
\label{app:intervention_flip}

\begin{center}\small
\captionof{table}{Prompt intervention TDR (\%) by flip type (Gemini, EN+JA).}
\label{tab:intervention_flip}
\begin{tabular}{lcccc}
\toprule
\textbf{Flip Type} & \textbf{base} & \textbf{adv} & \textbf{expl} & \textbf{aud-1st} \\
\midrule
adjective & 13.1 & 2.8 & 5.8 & 25.7 \\
number & 18.2 & 3.2 & 7.1 & 39.2 \\
time & 22.5 & 3.6 & 8.2 & 36.0 \\
neg\_add & 22.3 & 3.2 & 7.0 & 31.2 \\
neg\_remove & 21.5 & 8.9 & 15.2 & 31.0 \\
\bottomrule
\end{tabular}
\end{center}

\section{Error Rates}
\label{app:errors}

\begin{center}\small
\captionof{table}{Response parse validity in the conflict condition.}
\label{tab:errors}
\begin{tabular}{lccc}
\toprule
\textbf{Model} & \textbf{Valid} & \textbf{Parse Fail} & \textbf{Refusal} \\
\midrule
Gemini & 99.8\% & 0.1\% & 0.1\% \\
GPT-4o & 99.7\% & 0.2\% & 0.1\% \\
Ultravox & 99.7\% & 0.3\% & 0.0\% \\
Qwen2 & 98.9\% & 1.0\% & 0.1\% \\
\bottomrule
\end{tabular}
\end{center}

\section{Fine-Tuning Ablation Details}
\label{app:ablation}

\begin{center}\small
\captionof{table}{Ablation TDR (\%) by held-out flip type.}
\label{tab:ablation_fliptype}
\begin{tabular}{lccc}
\toprule
\textbf{Flip Type} & \textbf{Baseline} & \textbf{Adapter} & \textbf{LoRA} \\
\midrule
negation\_add & 59.3 & 80.8 & 32.4 \\
time\_swap & 48.5 & 75.9 & 21.7 \\
negation\_remove & 43.2 & 71.8 & 26.7 \\
\bottomrule
\end{tabular}

\bigskip

\captionof{table}{Ablation TDR (\%) by language (held-out flip types). LoRA dramatically reduces TDR for European languages but barely improves CJK/Arabic.}
\label{tab:ablation_language}
\begin{tabular}{lccc}
\toprule
\textbf{Lang} & \textbf{Baseline} & \textbf{Adapter} & \textbf{LoRA} \\
\midrule
EN & 37.4 & 77.7 & 7.6 \\
DE & 48.5 & 84.9 & 10.0 \\
FR & 50.1 & 84.5 & 12.1 \\
IT & 44.1 & 85.8 & 9.1 \\
PT & 48.0 & 82.2 & 13.9 \\
AR & 57.3 & 68.5 & 54.2 \\
JA & 52.8 & 53.3 & 48.5 \\
ZH & 56.9 & 69.4 & 49.8 \\
\bottomrule
\end{tabular}
\end{center}

\section{Example Stimuli}
\label{app:examples}

\begin{center}\small
\captionof{table}{Representative English stimuli, one per flip type. \textbf{Bold} marks the conflicting element.}
\label{tab:examples}
\resizebox{\columnwidth}{!}{%
\begin{tabular}{p{1.4cm}p{3.2cm}p{2.2cm}p{1.0cm}p{1.0cm}}
\toprule
\textbf{Flip Type} & \textbf{Audio / Text} & \textbf{Question} & \textbf{Audio} & \textbf{Text} \\
\midrule
number &
  ``...his \textbf{three} daughters'' \newline
  ``...his \textbf{two} daughters'' &
  How many daughters? & three & two \\
\midrule
neg\_add &
  ``Passengers \textbf{can} catch...'' \newline
  ``Passengers \textbf{cannot} catch...'' &
  Passengers allowed? & can & cannot \\
\midrule
adjective &
  ``a \textbf{new} skatepark'' \newline
  ``an \textbf{old} skatepark'' &
  What kind? & new & old \\
\midrule
time &
  ``...the \textbf{Friday} night'' \newline
  ``...the \textbf{Saturday} night'' &
  Which night? & Fri. & Sat. \\
\bottomrule
\end{tabular}}
\end{center}

\section{TTS Effect by Language}
\label{app:tts}

\begin{table*}[t]
\centering
\caption{Per-language TDR (\%) for natural vs.\ TTS audio. Format: Natural\,/\,TTS ($\Delta$).}
\label{tab:tts_lang}
\small
\begin{tabular}{lcccc}
\toprule
\textbf{Lang} & \textbf{Gemini} & \textbf{GPT-4o} & \textbf{Ultravox} & \textbf{Qwen2} \\
\midrule
EN & 8.1\,/\,6.7 ($-$1.4) & 11.3\,/\,7.4 ($-$4.0) & 39.2\,/\,43.1 (+3.8) & 69.2\,/\,77.1 (+7.8) \\
DE & 11.3\,/\,11.2 ($-$0.1) & 13.6\,/\,14.0 (+0.4) & 48.5\,/\,53.3 (+4.8) & 63.6\,/\,68.1 (+4.6) \\
FR & 12.6\,/\,14.8 (+2.2) & 19.8\,/\,17.2 ($-$2.6) & 47.7\,/\,54.0 (+6.3) & 71.0\,/\,73.3 (+2.3) \\
IT & 8.9\,/\,9.5 (+0.6) & 21.4\,/\,19.3 ($-$2.1) & 40.1\,/\,48.1 (+8.0) & 70.4\,/\,73.6 (+3.3) \\
PT & 10.3\,/\,13.0 (+2.7) & 21.9\,/\,20.0 ($-$1.9) & 49.4\,/\,53.0 (+3.6) & 71.5\,/\,73.0 (+1.5) \\
AR & 19.9\,/\,19.4 ($-$0.5) & 25.1\,/\,14.5 ($-$10.7) & 53.8\,/\,54.6 (+0.8) & 50.4\,/\,50.4 (+0.0) \\
JA & 30.2\,/\,28.3 ($-$1.9) & 30.2\,/\,24.1 ($-$6.1) & 55.1\,/\,55.6 (+0.5) & 51.8\,/\,52.2 (+0.3) \\
ZH & 31.8\,/\,32.4 (+0.6) & 42.8\,/\,35.7 ($-$7.1) & 57.6\,/\,57.0 ($-$0.6) & 56.2\,/\,54.0 ($-$2.2) \\
\midrule
\textbf{All} & 16.6\,/\,16.8 (+0.2) & 23.2\,/\,19.0 ($-$4.2) & 48.8\,/\,52.3 (+3.5) & 63.2\,/\,65.4 (+2.2) \\
\bottomrule
\end{tabular}
\end{table*}

\end{document}